\documentclass{article}

\usepackage{PRIMEarxiv}
\usepackage{xspace}
\usepackage[utf8]{inputenc} 
\usepackage[T1]{fontenc}    
\usepackage{hyperref}       
\usepackage{url}            
\usepackage{booktabs}       
\usepackage{amsfonts}       
\usepackage{nicefrac}       
\usepackage{microtype}      
\usepackage{lipsum}
\usepackage{fancyhdr}       
\usepackage{graphicx}       
\graphicspath{{media/}}     
\usepackage[section]{placeins}
\title{Prediction of Red Wine Quality Using One‐dimensional Convolutional Neural Networks
}

\author{
  Shengnan Di\\
  Faculty of Engineering and Food Science\\
  University of New South Wales\\
  Sydney NSW 2052\\
  \texttt{z5126835@zmail.unsw.edu.au},  \\
  \And
  Yang Yang \\
  Faculty of Engineering and Information Technology\\
  University of Technology Sydney \\
  Sydney NSW 2000\\
  \texttt{yang.yang-1@student.uts.edu.au}\\
}

\begin{document}
\maketitle

\begin{abstract}
As an alcoholic beverage, wine has remained prevalent for thousands of years, and the quality assessment of wines has been significant in wine production and trade. Scholars have proposed various deep learning and machine learning algorithms for wine quality prediction, such as Support vector machine (SVM), Random Forest (RF), K-nearest neighbors (KNN), Deep neural network (DNN), and Logistic regression (LR). However, these methods ignore the inner relationship between the physical and chemical properties of the wine, for example, the correlations between pH values, fixed acidity, citric acid, and so on. To fill the gap, this paper conducts the Pearson correlation analysis, PCA analysis, and Shapiro-Wilk test on those properties and incorporates 1D-CNN architecture to capture the correlations among neighboring features. In addition, it implemented dropout and batch normalization techniques to improve the robustness of the proposed model. Massive experiments have shown that our method can outperform baseline approaches in wine quality prediction. Moreover, ablation experiments also demonstrate the effectiveness of incorporating the 1-D CNN module, Dropout, and normalization techniques.
\end{abstract}

\keywords{Wine Quality \and Deep Learning \and CNN \and correlation analysis \and PCA}

\section{Introduction}
Despite the effect of the pandemic and extreme weather, the production of wine continues to grow. The United States, the world's largest consumer of wine, consumes 33 million hectoliters of wine each year. Italy, France, and Spain are the largest three wine-producing countries in the world, producing 20.8 million hectoliters, 20.2 million hectoliters, and 13.6 million hectoliters of wine in 2020. As a commodity that is distributed in large quantities, wine has a well-established industrial production process and distribution network. The identification and classification of wine quality is a very important part of modern food production and trade. The color of the wine, and especially the flavor and fragrance, define the quality and even the price of the wine\cite{waterhouse2016understanding}. As a complex product of grape juice fermentation, wine contains over 9,000 chemicals that play an important role in the organoleptic properties (appearance, aroma, and flavor) and other characteristics of wine. Wine contains about 97\% of water and ethanol, the remaining 3\% consists of organic acids, sugars, acetic acid, acetaldehyde, glycerol, higher alcohols, sorbitol and polyols, methanol, sulfites, amino acids, volatile esters, minerals, phenols \cite{waterhouse2016understanding}. These chemical compositions determine the organoleptic quality of the wine. The aroma of the wine is mainly derived from the volatile substances in the wine, such as 1-Hexanol, 3-Mercaptohexanol, etc. And the flavor of wine depends mainly on the phenolic substances in the wine \cite{waterhouse2016understanding}\cite{bhardwaj2022machine}.
The distribution of chemical components in the wine is influenced by many factors, such as the quality of the grapes, the weather conditions, and the geographical environment in which the grapes are grown, as well as the temperature, humidity, and duration of the fermentation process. Fluctuations in these chemicals can be observed by examining the physicochemical properties of the wine. These physicochemical properties like pH value, alcohol content, sulfur content, anthocyanin content, and other properties of wine can greatly reflect its quality \cite{cortez2009modeling}. It is important to note that these physicochemical characteristics do not independently affect the quality of the wine. Research on wine has revealed the interesting fact that there is an intrinsic link between the physicochemical features of the wine and that two or more features may work together to influence certain qualities of a wine. For example, the presence of ethanol can mask the acidity in the wine as well as lessen the concentration of some odors. The presence of organic acids lowers the pH of the wine and brings out the sourness. So, it can be seen that the content of ethanol and the content of organic acids, as well as the pH of the wine, have an impact on the presentation of sourness in the wine.\cite{waterhouse2016understanding}

Traditionally, wine quality assessment has relied on manual sensory evaluation. Experienced tasters look at the appearance, aromas and flavors of the wine in order to facilitate a comprehensive score. However, the disadvantages of this method are very obvious: first of all, manual sensory evaluation relies heavily on personal perception, and different tasters may give very different results. Second, due to the lack of standardized quality evaluation criteria, manual sensory evaluation can no longer be adapted to mass production. The experience of a taster comes from the analysis of data during his or her career. There is a very limited amount of data that one can remember and process. Unlike humans, artificial intelligence (AI) has the ability to remember and learn from massive amounts of data. After being trained by large amounts of data, AI can work regardless of subjective factors to adapt to modern massive production.

\section{Related Work}
As a popular research direction, there has been a lot of research on the application of machine learning to the quality inspection of wine. In 2021, Dahal’s team applied Support Vector Machine (SVM), Gradient Boosting Regressor (GBR), Ridge Regression (RR), and Deep Neural Network (DNN) on wine quality prediction and compared the results \cite{dahal2021prediction}. Another team applied the RF and Nave Bayes on wine quality forecasting and gained the accuracy of 65.83\% and 55.91\% respectively \cite{kumar2020red}.
And in 2020, Bipul Shaw studied the application of SVM, random forest and multilayer perceptron on wine quality prediction \cite{shaw2020wine}. KNN and RF were applied by Mahima and team members to predict wine quality\cite{gupta2020wine}. In 2021, Lia compared the results of the application of RF, decision tree, boosting algorithm, stochastic gradient descent, and SVC on wine quality prediction \cite{astutipredicting}. However, all of these studies failed to consider the inner relationship among the features of the wine. 

As we mentioned before, two or more physicochemical features of wine may work together to inflect the quality of the wine. Thus, in this research, we performed a Pearson correlation analysis of the physicochemical characteristics of the wines. The Pearson correlation analysis results in a number between -1 and 1. A positive number means that there is a positive correlation between two features, and a negative number means that there is a negative correlation between two features \cite{boslaugh2012statistics}. Convolutional neural networks (CNN) are often applied to deal with image classification problems due to their ability to capture the local spatial information which is presented by adjacent pixel values \cite{sun2020automatically}. The capability of CNN to capture inner relationship information among neighboring features may also work for wine quality prediction. Unlike other algorithms, CNNs do not flatten the sequential structure of the input information, so before we feed the physicochemical properties of the wine as input into the CNN, we can manage the order of the features. Placing features with strong correlations in close proximity can make the internal correlation between them easier to be captured by CNN. Rather than treating each characteristic in isolation, studying the internal relationships between characteristics may also be of great help in predicting the quality of a wine. In practice, we arranged the physical and chemical characteristics of wines with large positive or negative correlations next to each other and feed them into the CNN model. Moreover, for many previous studies, the limited sample size was a difficult problem to solve. It is difficult and expensive to perform physicochemical tests and obtain large amounts of data during the winemaking process\cite{bhardwaj2022machine}\cite{agrawal2018wine}. If the number of samples is not sufficient, it may lead to overfitting which can reduce the performance of the model. In this situation, another advantage of CNN which is ‘parameter sharing’ can solve this problem. In CNN, one filter is used to slide over the whole input dataset, which means the parameters will be ‘shared’ among all the inputs. So compared with the other algorithms such as DNN which need to arrange parameters for each input, CNN requires a relatively small number of parameters. Combining the above reasons, we chose 1D-CNN-based neural networks as the model to study in this paper. In deep learning, the dropout technique is another way to solve overfitting in the model. The previous study of Dahal’s team didn’t apply the dropout on their Deep Neural Network (DNN) model, so, in our study, we add dropout and normalization to the proposed model to improve the accuracy. To tackle the above challenges in wine quality prediction, this paper makes the following contributions:

\begin{itemize}
\item It conducts the Pearson correlation analysis, and PCA analysis to examine the physicochemical properties.
\item We incorporate 1D-CNN architecture to capture the correlations among neighboring features.
\item We implemented dropout and batch normalization techniques to improve the robustness of the proposed
model
\end{itemize}

\section{Data Analysis}

The wine quality prediction project aims to use the features of wine to predict its quality score. The quality score range from 0 to 10, which is correlated to a series of chemical and physical properties. Therefore, it is necessary to analysis and their relationship before feeding into the model. This Section represents an initial exploration of the dataset. This paper will examine 11 physiochemical properties of wine including fixed acidity ([tartaric acid] $g/dm3$), volatile acidity ([gacetic acid] $g/dm3$), total sulfur dioxide ($mg/dm3$), chlorides ([sodium chloride] $g/dm3$), pH level, free sulfur dioxide ($mg/dm3$), density ($g/cm3$), residual sugar ($g/dm3$), citric acid ($g/dm3$), sulfates ([potassium sulfate] $g/dm3$), and alcohol ($vol$\%). The dataset is available from the UCI machine learning repository  \url{https://archive.ics.uci.edu/ml/datasets/wine+quality}. It conducts data visualization, correlation analysis, PCA analysis, and Shapiro test in this Section.

\subsection{Pearson Correlation Analysis}

\begin{figure}[ht]
  \centering
  \includegraphics[width=0.45\textwidth]{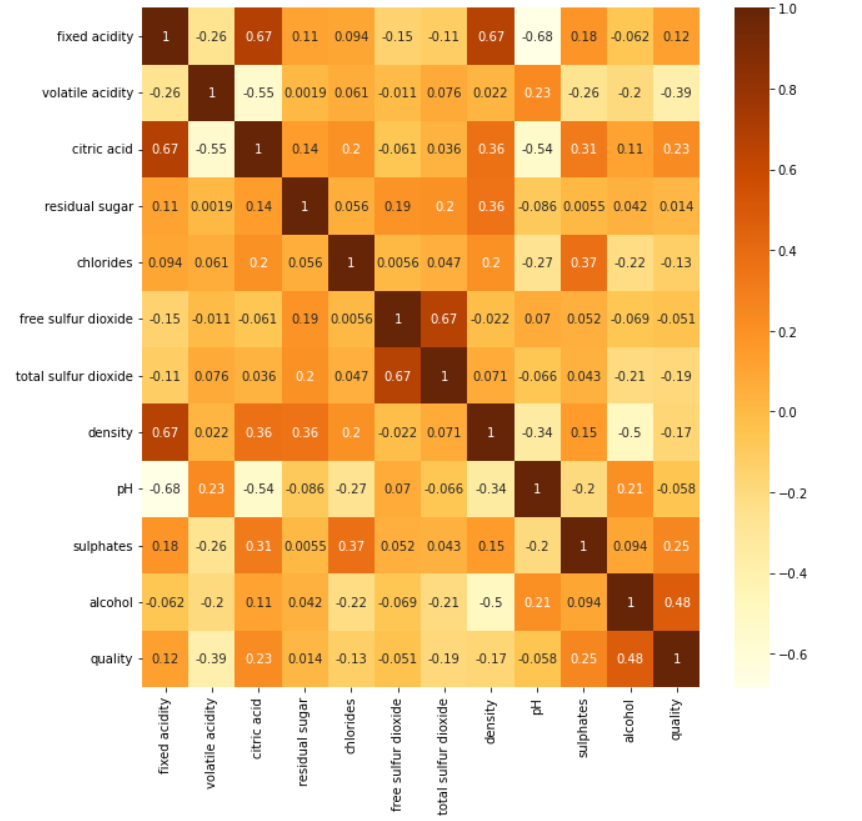}
  \label{fig:Pearson}
  \caption{Pearson Correlation Heatmap of Features}
\end{figure}

As the physicochemical properties can jointly affect the quality of wine and those properties are also correlated with each other, it is necessary to explore their relationship before the modeling process. Here, we implemented a statistical approach, the Pearson correlation coefficient (PCC)\cite{glen2021correlation}, to gauge the linear bivariate correlations of any two properties of the wine. Although red wine may have other kinds of correlations within its physicochemical characteristics, this approach can normalize the correlation coefficient into [-1, 1] and indicate the degree of linear correlation between any two features. Eq \ref{eq.PCC} describes the formula of the Pearson correlation coefficient.
And Figure \ref{fig:Pearson} is a heatmap of the Pearson correlation coefficient.

\begin{equation}
\rho_{X_{i}, X_{j}}=\frac{\mathbb{E}\left[\left(X_{I}-\mu_{X_{i}}\right)\left(X_{j}-\mu_{X_{j}}\right)\right]}{\sigma_{X_{i}} \sigma_{X_{j}}}
\label{eq.PCC}
\end{equation}

where $X_{i}$ and $X{j}$ denote a pair of features of the wine. $\mu$ and $\sigma$ represent the mean and standard deviation of the variables, respectively. Then,  the formula for the Pearson correlation coefficient $\rho$ can be expressed by Eq \ref{eq.PCC}. After calculating the value of $\rho$, we can obtain a heatmap as shown in Figure \ref{fig:Pearson}. The figure indicates that the highest linear correlation with the quality of red wine is the alcohol ($vol$\%) at 0.48, while the least correlated is the residual sugar ($g/dm3$) at 0.014. In addition, fixed acidity ([tartaric acid] $g/dm3$) and density ($g/cm3$), citric acid ($g/dm3$), and pH level also show a strong positive correlation.

\subsection{PCA Analysis}

Principal component analysis, known as PCA\cite{r2018r}, is a dimensionality reduction method frequently used to reduce the dimensionality of a large dataset by transforming a large set of variables into a smaller set of variables that still contains significant information in the large dataset.
A reduction in the number of variables in a dataset does affect the accuracy of the prediction, but the key point of dimensionality reduction is to trade a little accuracy for simplicity. This is because the transformed smaller datasets are easier to explore and visualize and allow machine learning algorithms to analyze the data more easily and faster without extraneous variable processing. In Figure \ref{tab:table3}, it utilized PCA decomposition to analyze the importance of the random variables.

\begin{table}[ht]
  \centering
  \resizebox{\textwidth}{7mm}{
  \begin{tabular}{lllllll}
    \cmidrule(r){1-7}
    Properties &  alcohol & sulphates & volatile
acidity & total sulfur dioxide & density & chlorides      \\
    \midrule
    Importance & 0.21 & 0.13 & 0.1 & 0.1 & 0.08 & 0.07 \\
    \bottomrule
  \end{tabular}}
  \setlength{\abovecaptionskip}{10pt}
  \label{tab:table}
\end{table}

\begin{table}[ht]
  \centering
  \resizebox{\textwidth}{7mm}{
  \begin{tabular}{lllllll}
    \cmidrule(r){1-6}
    Properties & pH level & fixed acidity & citric acid & free sulfur dioxide & residual sugar\\
    \midrule
    Importance & 0.06 & 0.06 & 0.06& 0.05 & 0.05\\
    \bottomrule
  \end{tabular}}
  \setlength{\abovecaptionskip}{10pt}
  \caption{Importance of the Wine Features}
  \label{tab:table3}
\end{table}

\subsection{Shapiro-Wilk Test}
We also conduct a Shapiro-Wilk Test\cite{shapiro1965analysis} on the features of wine. Shapiro-Wilk Test compares the sample distribution to a normal distribution in a statistically significant way to determine if the data show deviations from or are consistent with normality. The Shapiro-Wilk Test has two hypothesis assumptions. We need to calculate the $p$ value. If the test is non-significant, for example, $p<0.5$,  it indicates that the distribution of the sample is significantly different from a normal distribution. We calculate the $p$ values for each feature and the results show none of the variables follows a normal distribution.

\section{Methodology}

\subsection{Data Transformation}
To improve the learning efficiency and performance of the network, we normalize these physicochemical properties before feeding them into the neural network. Specifically, we implemented standard feature scaling to map each value into [0,1]. The Eq \ref{scaling} is the scaling process, where $\mu$ denotes the mean and $s$ is the standard deviation of the training samples.
\begin{equation}
z=\frac{x-\mu}{\sigma}
\label{scaling}
\end{equation}

\subsection{1D-CNN Architecture}

In the application of neural networks, if the task is a multi-class classification, we can choose the neural network with fewer hidden layers but with more hidden units in each layer, or the neural network with more hidden layers but fewer hidden units in each layer. A neural network with multiple layers of hidden layers is called a Deep Neural Network (DNN). Deep neural networks with many hidden layers can learn complex problems better. For deep neural networks, the early layers can learn simple features at lower levels, while the later layers can gather the simple information detected earlier for further analysis to solve the complex problem\cite{sze2017efficient}.

A convolutional neural network (CNN) is an algorithm that is mainly used in computer vision. In the field of computer vision (such as image classification and object detection), the objects that algorithms need to process are pictures. If the input image is a high-resolution clear image, then the image will be high in pixels, which means the dimension of the input X will be large. If we use a Deep neural network to process images, there will be a lot of hidden units. The computation will be expensive and that will also lead to overfitting. CNN can solve this problem. In the CNN model, there is a fixed-dimension filter that slides over the input matrix and performs convolution to obtain high-level features. Then followed by the activation function, dense layer, and output layer, you will get the result. One of the advantages of CNN is that the computational cost is lower than DNN because the parameter in the filter is shared among the input elements. The filter used in one part of the input image will also be used in another part of the image. This characteristic also prevents overfitting. Another advantage of CNN is the sparsity of the connections, in each layer of CNN, each output value depends only on a small number of neighboring inputs\cite{gu2018recent}.In this study, we use 1D-CNN on wine quality prediction to capture the inner correlation between the features and avoid overfitting. 

In this 1D-CNN network, first, the standard feature scaling was applied to the input data to scale the features. Then the order of the features was rearranged based on the results of Pearson Correlation Analysis to place features that are highly correlated with each other. The transformed and resorted features were fed into the convolutional layer to get the matrix which was then flattened to an array. Second, this flattened array went through 4 layers of deep neural networks followed by a softmax layer to get the prediction of the wine quality. The parameters were optimized by the backpropagation algorithm and the loss function is shown below:

\begin{equation}
Loss =-\sum_{i=1}^n y_i^* \log \hat{y}
\label{loss}
\end{equation}

It implements a Cross Entropy Loss for the back-propagation algorithms to update the whole learnable parameters of the 1D-CNN network. In our data, the wines are graded as integers between 1 and 10, so there are 10 categories in total and n equals 10. In Eq\ref{loss}, $y_{i}$ denotes the label data in the input dataset and $\hat{y}$ denoted the predicted label of wine quality.

\begin{figure}[ht]
  \centering
  \includegraphics[width=0.8\textwidth]{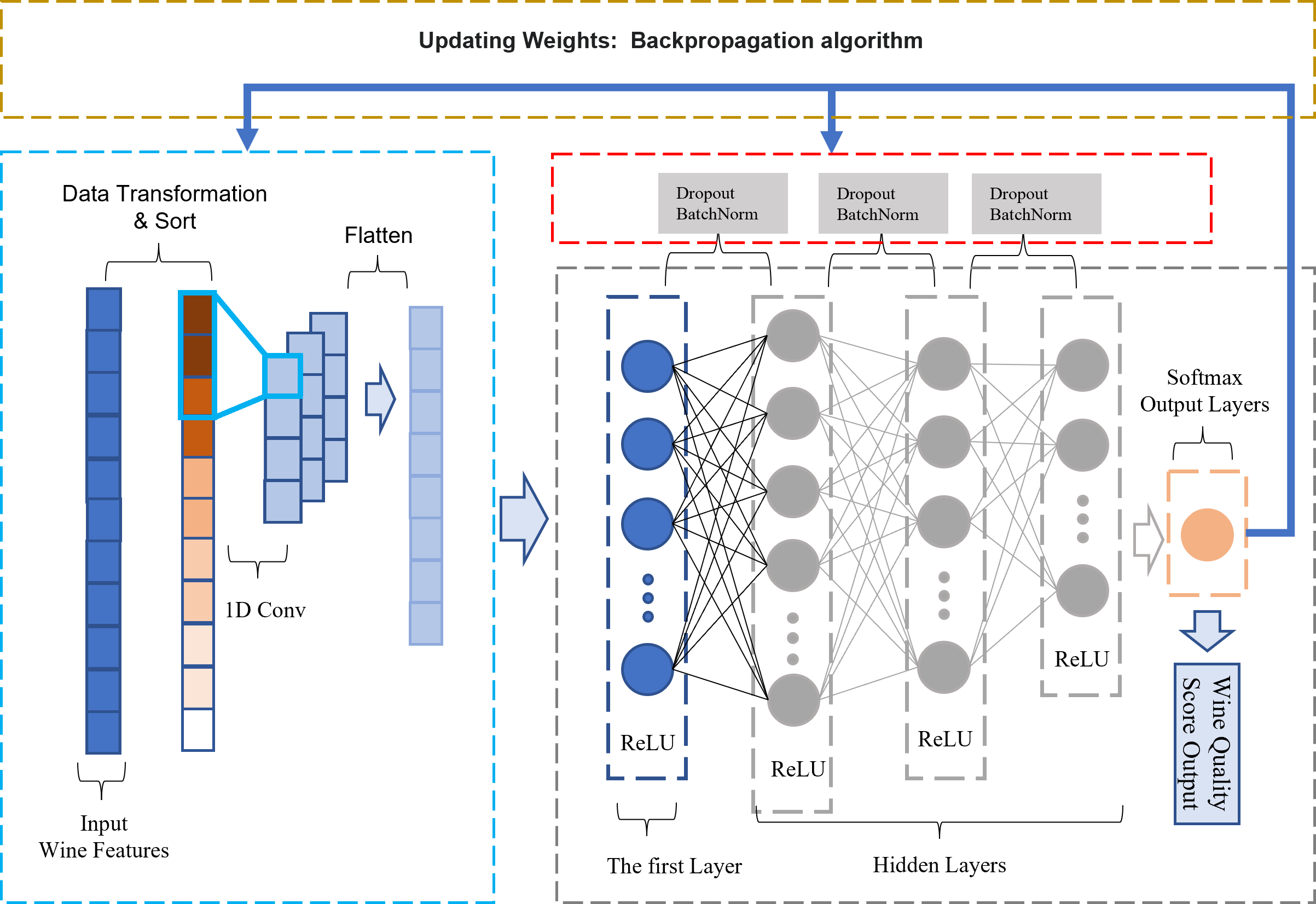}
  \label{fig:Architecture}
  \caption{1D-CNN Network Architecture}
\end{figure}

\subsection{Evaluation Metrics}
In this paper, we use statistical metrics to evaluate the performance of the model, including Recall, Precision, Accuracy, and F1 Score. These metrics are mainly used to compare the performance of classifiers. The higher these metrics indicate that the models have a better ability to perform the classification tasks. In particular, Recall is a metric that measures the amount of correct positive scores out of all possible positive scores made. While Precision only considers the correct positive predictions out of all predictions. Accuracy is also an important metric that is a measurement of all correctly identified cases. It is most commonly used when all classes are equally important. In addition, the F1 score is a better indicator of misclassification than the Accuracy metric as it is the harmonic mean of Precision and Recall.

\begin{equation}
Recall=\frac{TP}{TP+FN}
\end{equation}

\begin{equation}
Precision=\frac{T P}{T P+F P}
\end{equation}

\begin{equation}
Accuracy=\frac{T P+T N}{T P+T N+F P+F N}
\end{equation}

\begin{equation}
F1 Score=\frac{2(P * R)}{P+R}
\label{eq:F1}
\end{equation}

where the $TP$, $TN$, $FP$,and $FN$ denote True Positive, True Negative, False Positive, and False Negative, respectively. In Eq\ref{eq:F1}, $P$ and $R$ mean the values of Precision and Recall.

\section{Experimentation}
For the experiments, we first compare the proposed 1D-CNN network with a traditional machine learning baseline model including Support vector machine (SVM), Random Forest (RF), K-nearest neighbors (KNN), and Logistic regression (LR). Second, we conduct ablation experiments to evaluate the effectiveness of the modified blocks.

\subsection{Comparison Experiments: Setting and Baseline}
Table \ref{tab:table1} calculates the mean of selected metrics over ten classification tasks, aiming to give us an overall view of the model's performance. The results illustrate that the overall performance of the 1D-CNN network can outperform the baseline model. In particular, compared with the best baseline model, 1DCNN has a significant improvement where the mean value of Precision, Accuracy, Recall and F1 Score have increased at least 4\%, 2.7\%, 2.5\%, and 4.1\%, respectively.

\begin{table}[ht]
  \centering
  \begin{tabular}{lllll}
    \cmidrule(r){1-5}
    Model & Precision & Accuracy & Recall & F1-Score\\
    \midrule
    kNN & 0.761 & 0.776 & 0.761 & 0.768\\
    SVM & 0.752 & 0.765 & 0.754 & 0.760\\
    LR &  0.738 & 0.755 & 0.731 & 0.746\\
    RF &  0.813 & 0.810 & 0.825 & 0.824\\
    1DCNN & 0.849 & 0.832 & 0.846 & 0.858\\
    \bottomrule
  \end{tabular}
  \setlength{\abovecaptionskip}{10pt}
  \caption{Mean of Metrics for Wine Quality Classification}
  \label{tab:table1}
\end{table}

\subsection{Ablation Experiments: 1D-CNN Network}

To evaluate the effectiveness of the modified blocks including the 1D-CNN part and the Dropout layers. We design the ablation experiments to compare the performance in terms of Precision, Accuracy, Recall, and F1 Score. Specifically, we examine the following variants of the neural networks: 

\begin{itemize}
\item \textbf{DNN-D}: This model contains 4 fully connected layers only for the classification tasks.
\item \textbf{DNN}: This variant not only contains 4 fully connected layers but also has 3 Dropout layers and BatchNorm layers in the red dashed box of Figure\ref{fig:Architecture}.
\item \textbf{1DCNN-D}: This variant has 1DCNN block(the blue dashed box in Figure\ref{fig:Architecture})  but also does not use 3 Dropout layers and BatchNorm layers.
\item \textbf{1DCNN}: The 1D-CNN Network Architecture shown in Figure\ref{fig:Architecture}.
\end{itemize}

Table \ref{tab:table2} shows that the incorporation of the Dropout and BatchNorm layers can effectively improve the performance of the model. The models with that block can achieve a higher value in terms of Precision, Accuracy, Recall, and F1 Score with the enhancement of at least 1.4\%. Moreover, the combination of 1DCNN block can make the deep neural network have a better performance on Accuracy, Recall, and F1 Score. Last, from Table \ref{tab:table1} and Table \ref{tab:table2}, we find that the DNN-based models show a better overall performance on red wine quality prediction than traditional machine learning models such as kNN, SVM, LR, and RF.

\begin{table}[ht]
  \centering
  \begin{tabular}{lllll}
    \cmidrule(r){1-5}
    Model & Precision & Accuracy & Recall & F1-Score\\
    \midrule
    DNN-D   & 0.835 & 0.812 & 0.819 & 0.829\\
    DNN     & 0.851 & 0.825 & 0.834 & 0.850\\
    1DCNN-D &  0.836 & 0.820 & 0.823 & 0.824\\
    1DCNN   & 0.849 & 0.832 & 0.846 & 0.858\\
    \bottomrule
  \end{tabular}
  \setlength{\abovecaptionskip}{10pt}
  \caption{Ablation Experiments for Wine Quality Classification}
  \label{tab:table2}
\end{table}

\newpage
\section{Conclusion}
For modern food production and trade, quality forecasting of products is very important. There have been many studies on the application of artificial intelligence in the field of wine quality identification. However, researchers have encountered the problem of insufficient data when using deep learning to identify wine quality, which would result in overfitting. Moreover, none of the existing papers considers the impact of the intrinsic association between the physicochemical properties of the wine in the modeling process. To tackle the above challenges, we design 1D-CNN networks for the task of wine quality prediction. The 1D-CNN block can process adjacent features in one convolutional step. In addition, CNN requires fewer parameters and is less prone to overfitting problems even when the data is relatively small. We further design a Dropout block that includes 3 dropout layers and batch normalization layers to improve the robustness of the model and eliminate overfitting. The experiments and ablation study demonstrate the superiority and effectiveness of the proposed model.

However, exploring the quality of wine is complex due to the sophisticated correlations within its physicochemical properties. It is necessary for a model to consider the global relationship between those features and their interactions. In the future, it may need to explore more interaction between the features from a physicochemical perspective. For example, certain physical properties may be evolving along with the changes in multiple chemical compounds. In addition, there is also interconversion within chemical components. It remains an open problem to analyze the effect of different external variables, such as temperature, and light, on the quality of the wine.


\bibliographystyle{unsrt}  
\bibliography{references}

\begin{thebibliography}{10}

\bibitem{waterhouse2016understanding}
Andrew~L Waterhouse, Gavin~L Sacks, and David~W Jeffery.
\newblock {\em Understanding wine chemistry}.
\newblock John Wiley \& Sons, 2016.

\bibitem{bhardwaj2022machine}
Piyush Bhardwaj, Parul Tiwari, Kenneth Olejar~Jr, Wendy Parr, and Don Kulasiri.
\newblock A machine learning application in wine quality prediction.
\newblock {\em Machine Learning with Applications}, 8:100261, 2022.

\bibitem{cortez2009modeling}
Paulo Cortez, Ant{\'o}nio Cerdeira, Fernando Almeida, Telmo Matos, and Jos{\'e}
  Reis.
\newblock Modeling wine preferences by data mining from physicochemical
  properties.
\newblock {\em Decision support systems}, 47(4):547--553, 2009.

\bibitem{dahal2021prediction}
KR~Dahal, JN~Dahal, H~Banjade, and S~Gaire.
\newblock Prediction of wine quality using machine learning algorithms.
\newblock {\em Open Journal of Statistics}, 11(2):278--289, 2021.

\bibitem{kumar2020red}
Sunny Kumar, Kanika Agrawal, and Nelshan Mandan.
\newblock Red wine quality prediction using machine learning techniques.
\newblock In {\em 2020 International Conference on Computer Communication and
  Informatics (ICCCI)}, pages 1--6. IEEE, 2020.

\bibitem{shaw2020wine}
Bipul Shaw, Ankur~Kumar Suman, and Biswarup Chakraborty.
\newblock Wine quality analysis using machine learning.
\newblock In {\em Emerging technology in modelling and graphics}, pages
  239--247. Springer, 2020.

\bibitem{gupta2020wine}
Ujjawal Gupta, Yatindra Patidar, Abhishek Agarwal, Kushall~Pal Singh, et~al.
\newblock Wine quality analysis using machine learning algorithms.
\newblock In {\em Micro-Electronics and Telecommunication Engineering}, pages
  11--18. Springer, 2020.

\bibitem{astutipredicting}
Lia Astuti.
\newblock Predicting the important physicochemical quality of wine using the
  comparison of machine learning and deep learning.

\bibitem{boslaugh2012statistics}
Sarah Boslaugh.
\newblock {\em Statistics in a nutshell: A desktop quick reference}.
\newblock " O'Reilly Media, Inc.", 2012.

\bibitem{sun2020automatically}
Yanan Sun, Bing Xue, Mengjie Zhang, Gary~G Yen, and Jiancheng Lv.
\newblock Automatically designing cnn architectures using the genetic algorithm
  for image classification.
\newblock {\em IEEE transactions on cybernetics}, 50(9):3840--3854, 2020.

\bibitem{agrawal2018wine}
Garima Agrawal and Dae-Ki Kang.
\newblock Wine quality classification with multilayer perceptron.
\newblock {\em International Journal of Internet, Broadcasting and
  Communication}, 10(2):25--30, 2018.

\bibitem{glen2021correlation}
Stephanie Glen.
\newblock Correlation coefficient: Simple definition, formula, easy steps.
\newblock {\em StatisticsHowTo. com. Available online: https://www.
  statisticshowto.
  com/probability-and-statistics/correlation-coefficient-formula/(accessed on 3
  August 2020)}, 2021.

\bibitem{r2018r}
RFfSC R~Core~Team et~al.
\newblock R: A language and environment for statistical computing, 2018.

\bibitem{shapiro1965analysis}
Samuel~Sanford Shapiro and Martin~B Wilk.
\newblock An analysis of variance test for normality (complete samples).
\newblock {\em Biometrika}, 52(3/4):591--611, 1965.

\bibitem{sze2017efficient}
Vivienne Sze, Yu-Hsin Chen, Tien-Ju Yang, and Joel~S Emer.
\newblock Efficient processing of deep neural networks: A tutorial and survey.
\newblock {\em Proceedings of the IEEE}, 105(12):2295--2329, 2017.

\bibitem{gu2018recent}
Jiuxiang Gu, Zhenhua Wang, Jason Kuen, Lianyang Ma, Amir Shahroudy, Bing Shuai,
  Ting Liu, Xingxing Wang, Gang Wang, Jianfei Cai, et~al.
\newblock Recent advances in convolutional neural networks.
\newblock {\em Pattern recognition}, 77:354--377, 2018.

\end{thebibliography}

\end{document}